\documentclass[sigconf]{aamas} 
\usepackage{balance}  

\usepackage{booktabs}

\setcopyright{ifaamas}  
\acmDOI{} 
\acmISBN{} 
\acmConference[AAMAS'19]{Proc.\@ of the 18th International Conference on Autonomous Agents and Multiagent Systems (AAMAS 2019)}{May 13--17, 2019}{Montreal, Canada}{N.~Agmon, M.~E.~Taylor, E.~Elkind, M.~Veloso (eds.)}  
\acmYear{2019}  
\copyrightyear{2019} 
\acmPrice{} 

\usepackage{subfig}

\newcommand*\targ[1]{\overline{#1}}

\begin{document}

\title{Attention-based Deep Reinforcement Learning \\
for Multi-view Environments}  

\subtitle{Extended Abstract}

\author{Elaheh Barati}
\affiliation{%
  \institution{Wayne State University}
  \city{Detroit} 
  \state{MI} 
  \postcode{48202}
}
\email{elaheh.barati@wayne.edu}
\author{Xuewen Chen}
\affiliation{%
  \institution{AIWAYS AUTO}
  \city{Shanghai} 
  \country{China} 
}
\email{xuewen.chen@ai-ways.com}
\author{Zichun Zhong}
\affiliation{%
  \institution{Wayne State University}
  \city{Detroit} 
  \state{MI} 
  \postcode{48202}
}
\email{zichunzhong@wayne.edu}

\begin{abstract} 
In reinforcement learning algorithms, it is a common practice to account for only a single view of the environment to make the desired decisions; however, utilizing multiple views of the environment can help to promote the learning of complicated policies. Since the views may frequently suffer from partial observability, their provided observation can have different levels of importance. 
In this paper, we present a novel attention-based deep reinforcement learning method in a multi-view environment in which each view can provide various representative information about the environment. Specifically, our method learns a policy to dynamically attend to views of the environment based on their importance in the decision-making process. 
We evaluate the performance of our method on TORCS racing car simulator and three other complex 3D environments with obstacles. 
\end{abstract}


\maketitle

\section{Introduction} 
Distributed reinforcement learning algorithms~\cite{mnih2016asynchronous,barth2018distributed} have been proposed to improve the performance of the learning algorithm by passing copies of the environment to multiple workers. 
The adoption of multiple workers in these works is rather to increase the training reward and earlier convergence, not to collectively increase the amount of observable information from the environment through multiple sensory inputs. 

In a realistic environment, observability can be typically partial on account of occlusion from the obstacles or noise that affect the sensors such as cameras in the environment. 
Utilizing only one camera view can result in failure, since locating a single camera in a position that can capture both the targets as well as details of agents body is difficult~\cite{tassa2018deepmind}. On the other hands, sensory inputs with less importance can sometimes provide observations which are vital for achieving rich behavior. 
Therefore, it is desirable to incorporate multiple sensory inputs in the decision-making process according to their importance and their provided information at each time step. 
It reduces the sensitivity of policies to an individual sensor and makes the system capable of functioning despite one or more sensors malfunctioning.
Since sensors can provide diverse views of the environment and they are likely to be perturbed by different noise impacts, a policy is required to attend to the views accordingly.

\begin{figure}[!t]
 \centering
 \includegraphics[width=0.9\columnwidth]{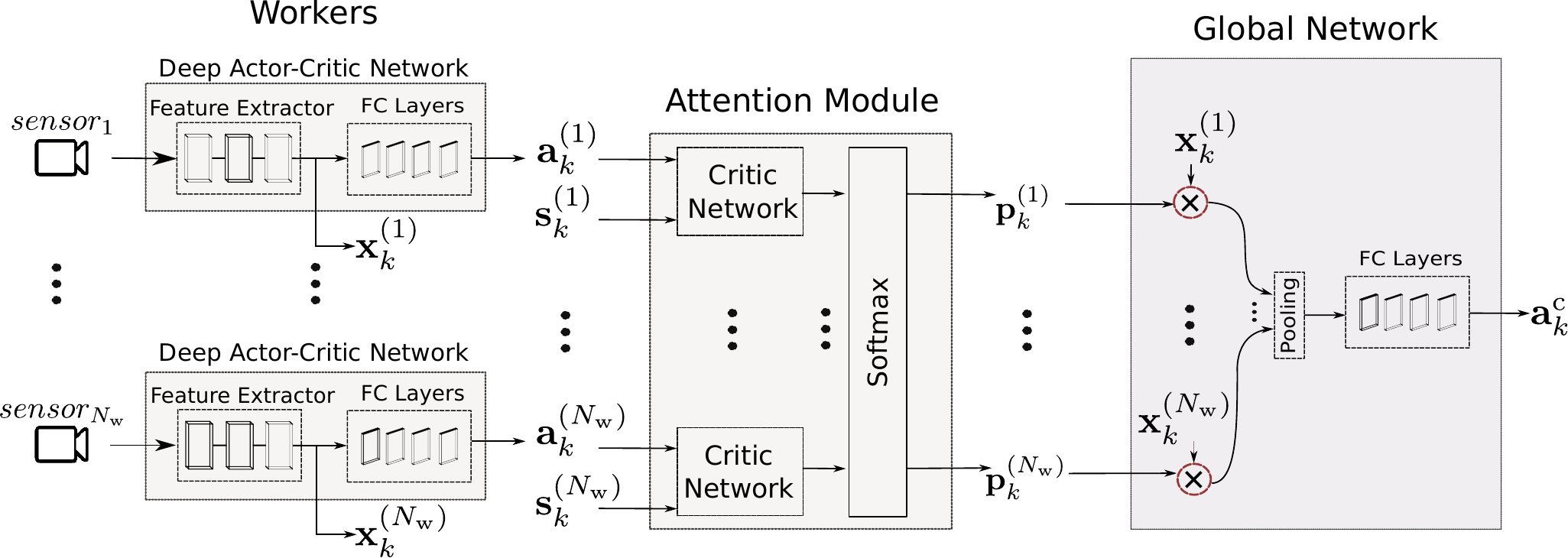}
 \caption{Architecture of the deep network that leverages attention mechanism in its global network. ${\bf p}^{(w)}_k$ is the weight of worker $w$ obtained from the attention module according to the importance of its view.}\label{fig:collaborative-model}
\end{figure}

In this paper, we propose an attention-based deep reinforcement learning method (depicted in Figure~\ref{fig:collaborative-model}) that learns a policy to attend to different views of the environment based on their importance. 
Each sensory input, which provides a specific view of the environment, is assigned to a worker. 
We employ an extension of the actor-critic approach~\cite{silver2014deterministic} to train the network of each worker and make the final decision through the integration of feature representations provided by the workers using an attention mechanism.
Since the critic network of each worker provides a signal regarding the amount of salient information supplied by its corresponding view, we employ this signal in the attention module to estimate the amount of impact that each of the views should have in the final decision-making process.
\section{Attention-based RL Framework}

\begin{figure*}[!t]
 \centering
 \begin{tabular}{l}
  \subfloat[TORCS]
  {\includegraphics[width=.52\columnwidth]{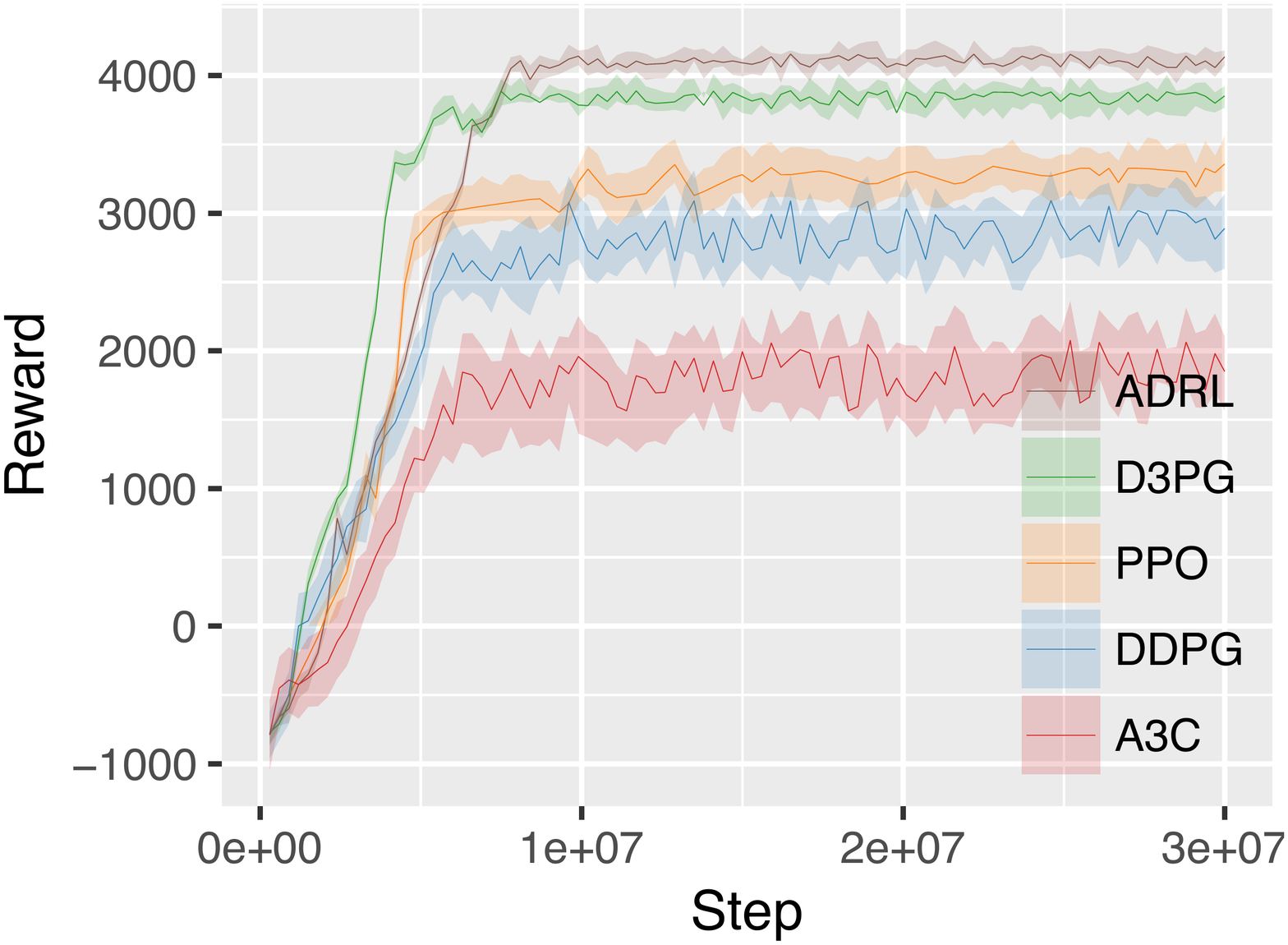}}
  \subfloat[Ant-Maze]{\includegraphics[width=.52\columnwidth]{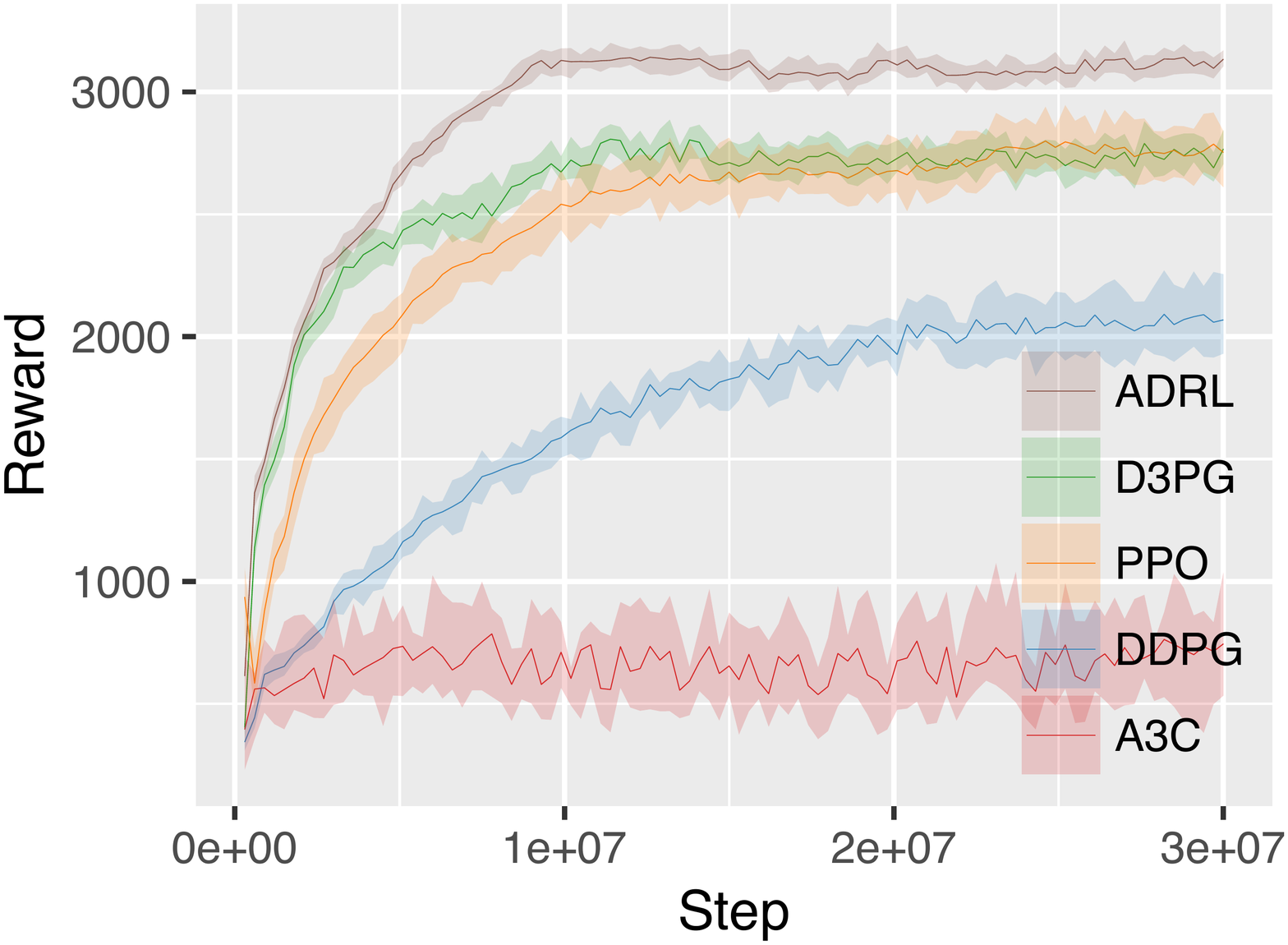}}
  \subfloat[Hopper-Stairs]
  {\includegraphics[width=.52 \columnwidth]{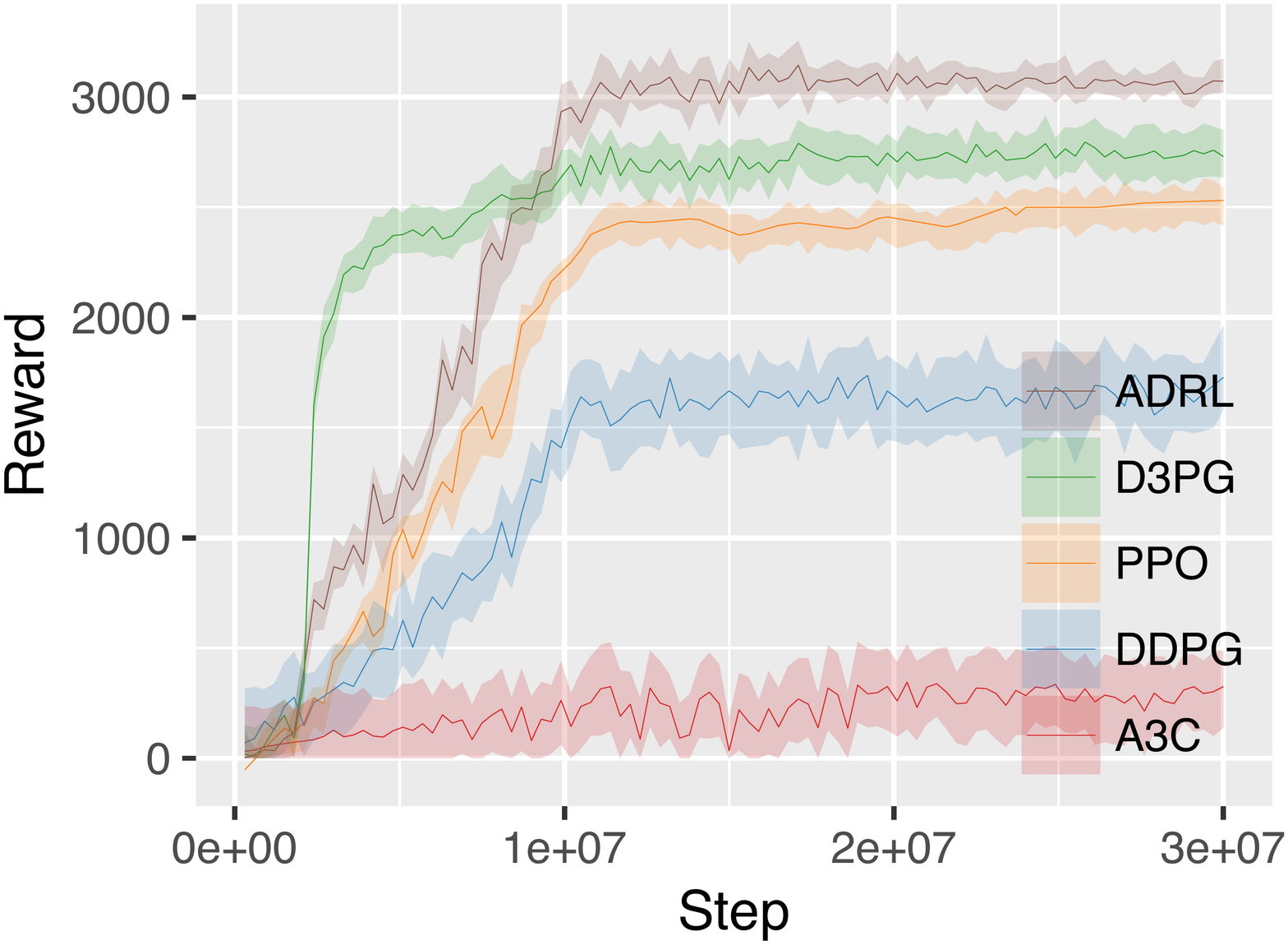}}
  \subfloat[Walker-Wall]{\includegraphics[width=.52\columnwidth]{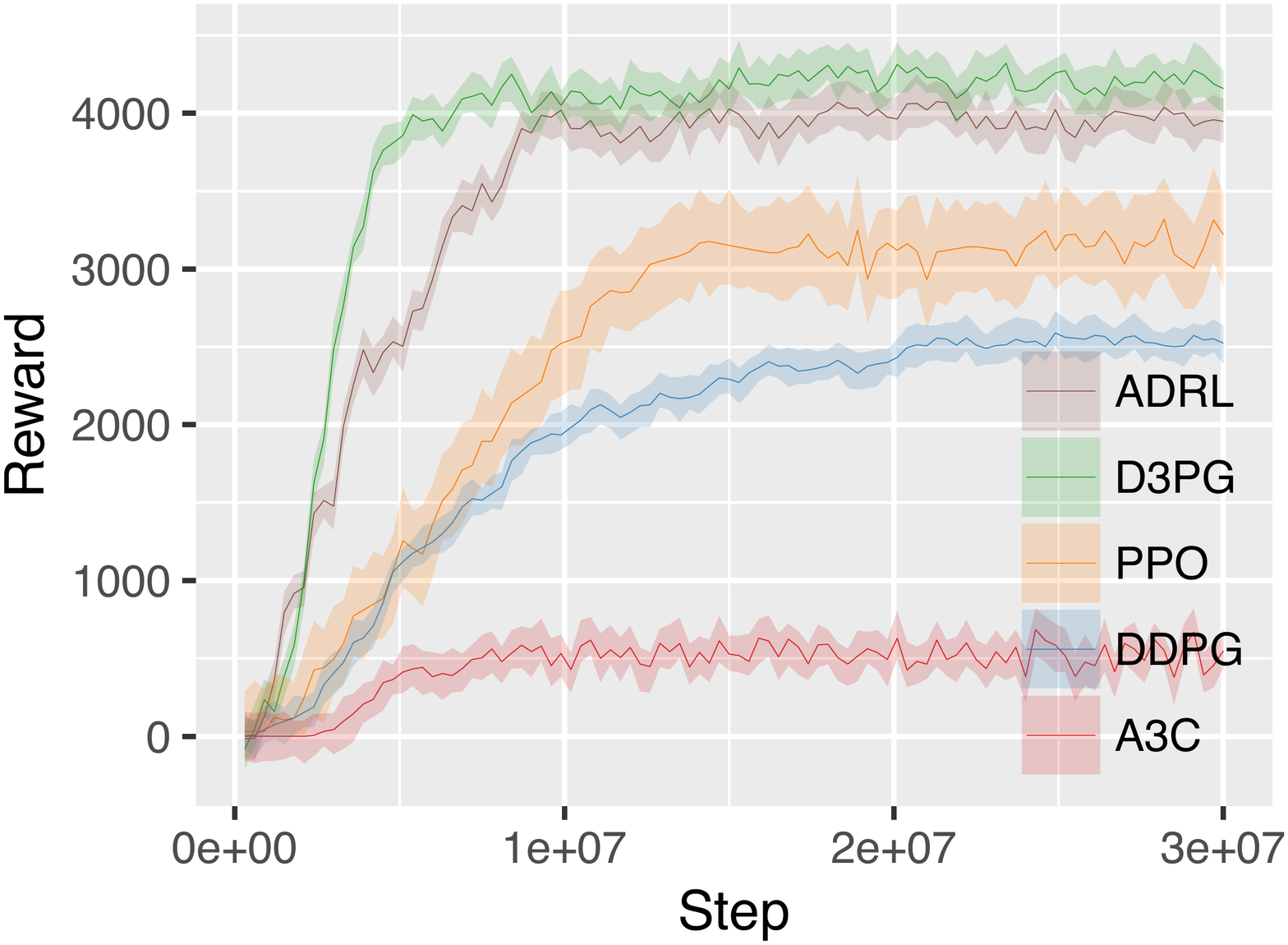}} 
 \end{tabular}
 \caption{Average reward vs.\ training
  step for the methods \texttt{DDPG}, \texttt{D3PG}, \texttt{PPO}, 
  \texttt{A3C}, and \texttt{ADRL}. We obtain the rewards in these figures by averaging rewards obtained from 5 runs. }\label{fig:training}
\end{figure*}

By utilizing an attention weighted representation as introduced in~\cite{bahdanau2014neural}, we incorporate the importance of the views in computing a unit representation of the environment (${\bf x}_k$). We use a softmax gate function to learn the attention module as
\begin{equation}\label{eq:pool}
 {\bf x}_k = \sum_{w=1}^{N_{\rm w}}{{\bf p}^{(w)}_k \odot {\bf x}^{(w)}_k} = \sum_{w=1}^{N_{\rm w}}{\frac{\exp(g_w f_w)}{\sum_{l}{\exp(g_lf_l)}} \odot {\bf x}^{(w)}_k}\; ,
\end{equation}
where $\odot$ stands for Hadamard product, ${\bf x}^{(w)}_k$ is the feature representation obtained by worker $w$, and $g_w$ is a parameter of the model.
In~\eqref{eq:pool}, $f_w$ is obtained from the output of critic network learned for each worker $w$ separately, i.e.,
$f_w = Q(s^{(w)}, a^{(w)})\;.$
At each time step $k$, the attention mechanism generates a positive weight ${\bf p}^{(w)}_k$ for each worker which determines the relative importance of worker $w$ in blending the feature vectors $\{{\bf x}^{(w)}_k| w=1,2,\ldots,N_{\rm w}\}$.

During training of our model, the aim is to promote the behavior of each worker by comparing its selected action with the actions selected by all the other workers.
To do so, we modify the reward at training of the global network by introducing a penalty term that depends on the actions selected by all the workers ($A_k$) as:
\begin{align}\label{eq:r_c}
 r^{\rm c}_k &= r_k - \gamma_{\rm r}{\delta(A_k)} =r_k - \gamma_{\rm
 r}\frac{1}{N_{\rm w}} \sum_{w=1}^{N_{\rm w}} \delta^{(w)}(A_k)
\end{align}
where $r_k$ is the original reward, $\gamma_{\rm r}$ is a constant that provides a trade-off between the deviation of actions and the original reward, and $A_k$ is the action matrix with columns ${\bf a}^{(w)}_k$. In~\eqref{eq:r_c}, we define
\begin{align}\label{eq:delta0}
\delta^{(w)}(A_k) = \bigg\vert\bigg\vert {\bf a}^{(w)}_k - \frac{1}{N_{\rm w}-1}\sum_{\substack{v=1\\v\neq w}}^{N_{\rm w}} {\bf a}^{(v)}_k\bigg\vert\bigg\vert^2 \;
\end{align}
as a deviation function that depends on the variation of action ${\bf a}^{(w)}_k$ of worker $w$ from the average of actions selected by the other workers in the network.
In~\eqref{eq:delta0}, the first term is the action of worker $w$ while the second term is the average action of other workers. 
In the case that all workers have trained sufficiently, the deviation $\delta(A_k)$ becomes close to zero, while by experiencing weak training performance from a worker, we get a higher $\delta(A_k)$ and a lower reward $r^{\rm c}_k$. Therefore, the penalty term in~\eqref{eq:r_c} enforces improvement in the training of the workers that are yet to be trained sufficiently.
We consider $\gamma_{\rm r}$ in~\eqref{eq:r_c} to be $0.1$ in our experiments.

The final action (${\bf a}^{\rm c}_k$) is determined by the global network from an aggregation of feature representations (${\bf s}^{\rm c}_k = {\bf x}_k$) obtained from the workers. 
To train the parameters of global network ($\theta^{\mu^{\rm c}}_k$, $\theta^{Q^{\rm c}}_k$), at each time step $k$, we stack the modified reward $r^{\rm c}_k$ on the replay buffer, forming a tuple ${\prec} {\bf s}^{\rm c}_k, {\bf a}^{\rm c}_k, r^{\rm c}_k,
{\bf s}_{k+1}^{\rm c}{\succ}$. Then, we sample a random mini-batch from the replay buffer to train critic network ($\theta^{Q^{\rm c}}_k$) by minimizing the following loss function~\cite{lillicrap2015continuous}:
\begin{equation*}
 {
  {\mathcal{L}}(\theta^{Q^{\rm c}}_k) = {\mathbb{E}}\bigg(
  r^{\rm c}_k +
  \gamma \overline {Q}({\bf s}^{\rm c}_{k+1},\targ\mu^{\rm c}({\bf s}^{\rm c}_{k+1};\theta^{\targ\mu^{\rm c}});\theta^{\targ Q^{\rm c}})-
  Q({\bf s}^{\rm c}_k,{\bf a}^{\rm c}_k;\theta^{Q^{\rm c}}_k) \bigg)^2
 }\;.
\end{equation*}
We update the actor network at each time step with respect to the set of parameters $\theta^{\mu^{\rm c}}_k$ by using sampled policy gradient:
\begin{equation}
 {
  \nabla_{\theta^{\mu^{\rm c}_k}}J = {\mathbb{E}}\bigg(\nabla_{\bf a} Q({\bf s},{\bf a};\theta^{Q^{\rm c}}_k)|_{{\bf s}={\bf s}^{\rm c}_k , {\bf a}=\mu^{\rm c}({\bf s}^{\rm c}_k)}
  \nabla_{\theta_\mu^{\rm c}}\mu^{\rm c}({\bf s};\theta^{\mu^{\rm c}}_k)|_{{\bf s}={\bf s}^{\rm c}_k}\bigg)\; .
 }
\end{equation}

\section{Experiments}
We compare the performance of our method with the baselines on tasks, Ant-Maze, Hopper-Stairs, and Walker-Wall developed in the MuJoCo physics simulator~\cite{todorov2012mujoco}. We also verify our method for autonomous driving on an open-source platform for car racing called TORCS~\cite{wymann2000torcs}.
The baselines \texttt{D3PG}~\cite{barth2018distributed}, \texttt{PPO}~\cite{schulman2017proximal}, \texttt{DDPG}~\cite{lillicrap2015continuous}, and \texttt{A3C}~\cite{mnih2016asynchronous} are state-of-the-art actor-critic based methods that are designed to work on continuous action spaces.
Figure~\ref{fig:training} shows the training speed of our method, Attention-based Deep RL (\texttt{ADRL}), and four of its baselines in terms of average reward per step. 
At the first stage in training of \texttt{ADRL}, multiple workers are trained separately given multiple views of the environment, and the reward of \texttt{ADRL} is the average reward of all these workers. 
Since \texttt{D3PG} uses all its workers to train the network given a single view of the environment, at the initial steps of training, the reward of \texttt{ADRL} is less than \texttt{D3PG} for all the tasks other than Ant-Maze. However, after convergence, the reward of \texttt{ADRL} is either higher or comparable to \texttt{D3PG} in all the tasks which is an indication of the advantage of attending to multiple various views of the environment. 
In Ant-Maze task, different camera views provide significantly diverse information about the environment, and leveraging these views via the attention mechanism in \texttt{ADRL} leads to a significantly higher reward for \texttt{ADRL} in comparison to its baselines.
In Walker-Wall task, the reward of \texttt{ADRL} is slightly less than \texttt{D3PG} which is due to having a low diversity between the camera views provided from the environment.

\section{Conclusion}
Our method, \texttt{ADRL}, takes advantage of multiple views of the environment to obtain a stabilized training policy. \texttt{ADRL} dynamically attends to views according to their importance in the final decision-making process. To measure the importance of each view, \texttt{ADRL} uses the output of the critic network designated for that view. Through the experiments, we observed that \texttt{ADRL} outperforms its baselines.


\bibliographystyle{ACM-Reference-Format}  
\balance 
\bibliography{main}

\end{document}